\documentclass[letterpaper, 10 pt, conference]{ieeeconf}  

\IEEEoverridecommandlockouts                              

\usepackage{graphics} 
\usepackage{multirow} 
\usepackage{amsmath,amsfonts}
\usepackage{algorithmicx}
\usepackage{algpseudocode}
\usepackage{array}
\usepackage[caption=false,font=footnotesize,labelfont=rm,textfont=rm]{subfig}
\usepackage{textcomp}
\usepackage{stfloats}
\usepackage{url}
\usepackage{verbatim}
\usepackage{graphicx}
\usepackage[noadjust]{cite}
\usepackage{amsmath} 
\usepackage{amssymb}  
\usepackage[ruled,linesnumbered]{algorithm2e}
\usepackage{subfig}
\title{Predictive Kinematic Coordinate Control for Aerial Manipulators based on Modified Kinematics Learning}
 
\author{Zhengzhen Li$^{1,2}$, Jiahao Shen $^{2}$, Mengyu Ji$^{2}$, Huazi Cao$^{2,3}$, Shiyu Zhao$^{2}$
\thanks{This work was supported by National Major Research \& Development Plan - Intelligent Robotics Major Special Project (2023YFB4705500) and the Research Center for Industries of the Future at Westlake University (WU2022C027) }
\thanks{Corresponding author: Huazi Cao}
\thanks{$^{1}$College of Computer Science and Technology, Zhejiang University, Hangzhou, China. $^{2}$WINDY Lab, Department of Artificial Intelligence, Westlake University, Hangzhou, China. $^{3}$Westlake Institute for Optoelectronics, Hangzhou, China. \{lizhengzhen, shenjiahao, jimengyu, caohuazi, zhaoshiyu\}@westlake.edu.cn}
}

\begin{document}

\newcommand{\I}{\mathbf{I}}

\renewcommand{\O}{\Omega}
\newcommand{\Obar}{\bar{\Omega}}
\newcommand{\sat}{\mathrm{sat}}
\renewcommand{\d}{\mathrm{d}}

\newcommand{\blue}{\textcolor{blue}}
\newcommand{\red}{\textcolor{red}}

\newcommand{\A}{\mathbf{A}}
\renewcommand{\a}{\mathbf{a}}
\newcommand{\B}{\mathbf{B}}
\newcommand{\e}{\mathbf{e}}
\newcommand{\F}{\mathbf{F}}

\newcommand{\g}{\mathbf{g}}
\newcommand{\G}{\mathcal{G}}
\renewcommand{\H}{\mathbf{H}}
\newcommand{\h}{\mathbf{h}}
\newcommand{\E}{\mathbb{E}}
\newcommand{\J}{\mathcal{J}}

\newcommand{\M}{\mathbf{M}}
\newcommand{\N}{\mathcal{N}}

\renewcommand{\P}{\mathbf{P}}
\newcommand{\p}{\mathbf{p}}
\newcommand{\R}{\mathbf{R}}
\renewcommand{\S}{\mathbf{S}}
\newcommand{\s}{\mathbf{s}}

\newcommand{\U}{\mathbf{U}}
\newcommand{\smallu}{\mathbf{u}}
\newcommand{\V}{\mathcal{V}}
\newcommand{\vel}{\mathbf{v}}
\newcommand{\Vel}{\mathbf{V}}
\newcommand{\q}{\mathbf{q}}
\newcommand{\f}{\mathbf{f}}
\newcommand{\bk}{\mathbf{K}}

\newcommand{\w}{\mathbf{w}}
\newcommand{\W}{\mathbf{W}}
\newcommand{\x}{\mathbf{x}}
\newcommand{\y}{\mathbf{y}}
\newcommand{\z}{\mathbf{z}}

\newcommand{\zero}{\mathbf{0}}

\newcommand{\Eta}{\boldsymbol{\eta}}
\newcommand{\bomega}{\boldsymbol{\omega}}
\newcommand{\btau}{\boldsymbol{\tau}}

\graphicspath{{figures/}}

\maketitle
\thispagestyle{empty}
\pagestyle{empty}

\begin{abstract}
High-precision manipulation has always been a developmental goal for aerial manipulators. This paper investigates the kinematic coordinate control issue in aerial manipulators. We propose a predictive kinematic coordinate control method, which includes a learning-based modified kinematic model and a model predictive control (MPC) scheme based on weight allocation. Compared to existing methods, our proposed approach offers several attractive features. First, the kinematic model incorporates closed-loop dynamics characteristics and online residual learning. Compared to methods that do not consider closed-loop dynamics and residuals, our proposed method has improved accuracy by 59.6$\%$. Second, a MPC scheme that considers weight allocation has been proposed, which can coordinate the motion strategies of quadcopters and manipulators. Compared to methods that do not consider weight allocation, the proposed method can meet the requirements of more tasks. The proposed approach is verified through complex trajectory tracking and moving target tracking experiments. The results validate the effectiveness of the proposed method.
\end{abstract}
\section{introduction}

Aerial manipulators have received considerable attention in recent years \cite{ollero2021past,staub2018towards}. An increasing number of research efforts are employing aerial manipulators for tasks unachievable by quadcopters alone. This is primarily due to their ability to combine the rapid mobility of quadcopters with the high-precision manipulation capabilities of manipulators. Compared to quadcopters, aerial manipulators can interact with the environment to accomplish a variety of tasks, such as pick-and-place \cite{wang2023millimeter, luo2023time}, 3D construction \cite{zhang2022aerial}, and opening/closing doors \cite{orsag2014valve}. Relative to ground-based manipulators, aerial manipulators can access areas that are challenging for humans or ground robots to reach, significantly expanding the scope of manipulation tasks.

Achieving high-precision control has always been a goal in the research of aerial manipulators. Existing studies achieve high-precision control primarily through two aspects: dynamic control and kinematic control. Dynamic control refers to designing control laws for the quadcopter and the manipulator based on the dynamic model of the aerial manipulator, thereby offsetting the effects of dynamic coupling on the aerial manipulator. Dynamic control methods can be categorized into two types: the first is decoupled control \cite{cao2023eso, liu2021ddpg, lee2020aerial, kim2017robust}, where controllers are designed separately for the quadcopter and the manipulator. The second type is coupled control, where a nonlinear model of the entire aerial manipulation system is established to achieve coordinated control \cite{yang2014dynamics, yuksel2016differential}. However, starting from dynamic control, only centimeter-level control precision can be obtained because obtaining an accurate dynamic model of aerial manipulator is challenging. To achieve higher precision, research on kinematic control is also essential.

Kinematic control starts from the kinematics of aerial manipulators, focusing on achieving high-precision end-effector control through coordinating the movements of the quadcopter and the manipulator. Kinematic control primarily involves managing the position, velocity, or acceleration within the control state space to achieve precise trajectory tracking of the end-effector. Research in kinematic control primarily addresses two main aspects of these issues.

The first issue is how to achieve high-precision control. Current kinematic control strategies mainly fall into two categories: The first type is closed loop inverse kinematics control (CLIK), which adjusts the kinematic control strategy based on real-time end-effector error, thus achieving high-precision kinematic control \cite{yang2014dynamics, cataldi2019set}. The second method is optimization control, which enhances precision by selecting the optimal control strategy \cite{lunni2017nonlinear}. Compared to CLIK, it ensures strict adherence to physical constraints. An accurate model is a prerequisite for high-precision kinematic planning. However, due to factors such as coupling disturbances between the manipulator and the quadcopter, obtaining an accurate kinematic model of the aerial manipulator is quite challenging.

The second issue is how to allocate and switch control modes. Since the aerial manipulator consists of a quadcopter and a manipulator, its operational modes can be divided into flight mode, coordinated manipulation mode, hover manipulation mode, and configuration adjustment mode.  Different modes are adapted to different tasks, making it crucial to select the appropriate mode for specific tasks. Motion allocation can be achieved by adjusting the weights in the kinematic Jacobian matrix \cite{xing2020enhancement}. However, it cannot account for physical constraints, which can easily lead to system instability and safety issues. Another common method is optimization control, which ensures high manipulability of the manipulator while minimizing the coupling disturbances it imparts to the quadcopter \cite{geisert2016trajectory, imanberdiyev2020redundancy} , all under the premise of fulfilling the task. However, fixed weights may not adequately adapt to changes in the environment or to the requirements of different tasks. 

In response to the existing issues of insufficient precision and difficulties in motion allocation, we propose the predictive kinematic coordinate control method. The output of the kinematic controller, after undergoing double integration, serves as the input to the lower level decoupled dynamic controller. The quadcopter employs an extended state observer (ESO) based robust controller in our previous work \cite{cao2023eso}, while the manipulator utilizes a PID controller. Our predictive kinematic coordinate control method is comprised of a learning-based modified kinematic model and a model predictive control (MPC) scheme based on weight allocation, which enables high-precision end-effector kinematic control and coordinated motion allocation. The main novelties of our approach are as follows:

1) We propose a learning-based modified kinematic model for aerial manipulator high precision kinematic control. In the modified kinematic model, the system's dynamic characteristics are considered through an equivalent model, enhancing the precision of kinematic control. Compared to using traditional integral kinematic controller, the average error with the modified kinematic model controller has been reduced by 59.6 $\%$.

2)  An MPC approach based on weight allocation has been proposed. Given that the manipulator plays a dominant role during the operational process of the aerial manipulator, we evaluate the mobility of the manipulator and adjust the weights in the MPC objective function to coordinate the motion strategies of both the quadcopter and the manipulator, achieving coordinated control. Compared to methods that do not consider weight allocation, our proposed approach can adapt to a wider range of scenarios and tasks, thereby enhancing the applicability of aerial manipulators.

\begin{figure*}
\centering
\includegraphics[width=0.9\linewidth]{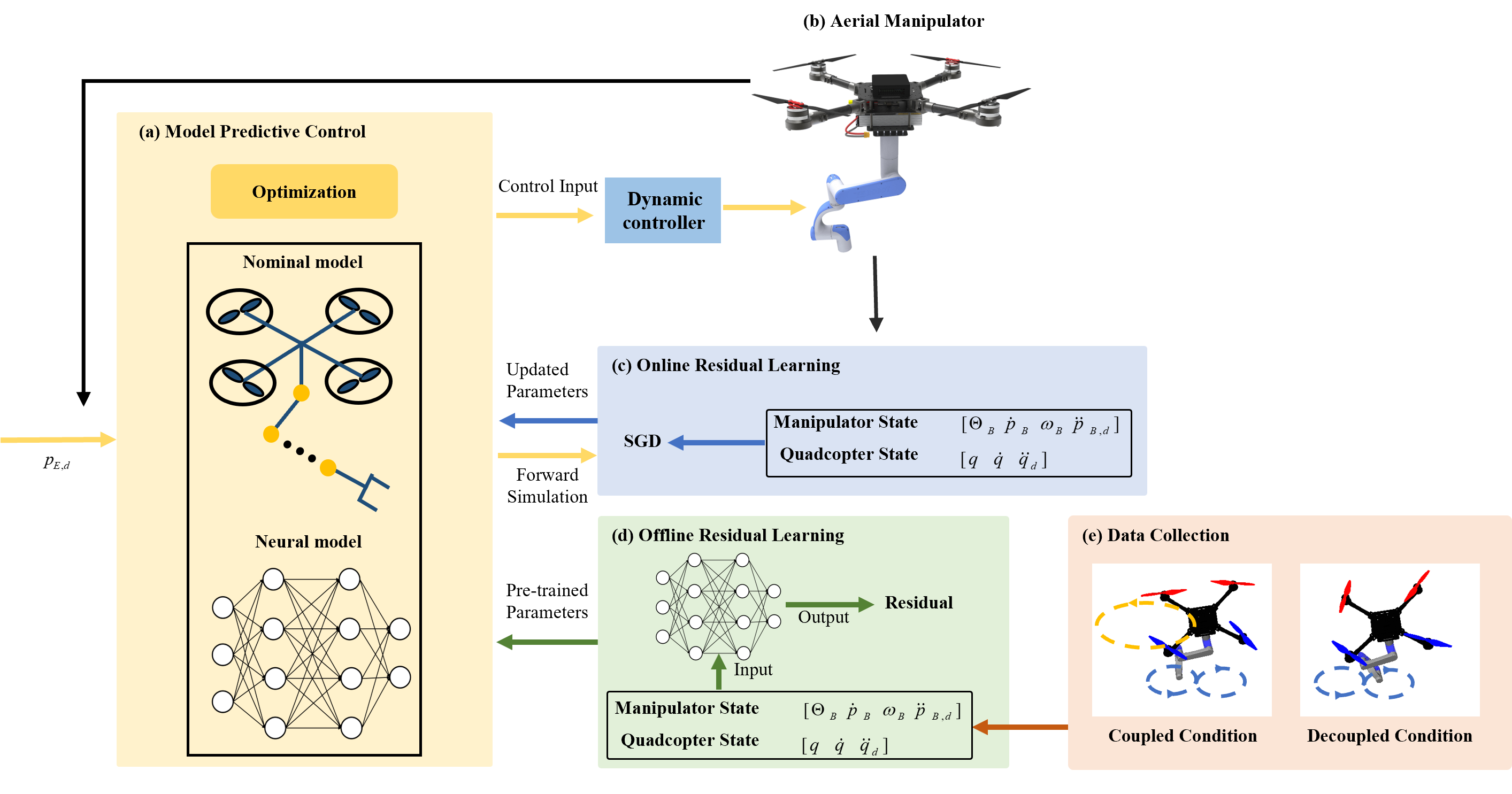}
\caption{\label{frame}Schematic of the MPC with learning based modified kinematic model.}
\end{figure*}

\section{Learing-based modified kinematic model}

This section describes the learning-based modified kinematic model used for coordinated kinematic control. Our modified kinematic model consists of two parts: the equivalent kinematic model of the quadcopter and the kinematic model of the manipulator. 

\subsection{Kinematics of Aerial Manipulator}

Let $\mathbf{R}_B \in SO(3)$ denote the rotation matrix of the quadcopter, while \(\mathbf{\Theta}_B = [\phi_B, \theta_B, \psi_B] \in \mathbb{R}^3\)  represents the attitude angle vector of the quadcopter. The quadcopter's position and the end-effector position are denoted by $\mathbf{p}_B, \mathbf{p}_E \in \mathbb{R}^3$, respectively.  Let \(n\) denote the number of joints in the manipulator,  \(\mathbf{q} = [q_1, q_2, ..., q_n]^\mathrm{T} \in \mathbb{R}^n\) denote the joint angle vector of the manipulator, and $n=6$ in this work. The end-effector position in $\Sigma_B$ is denoted by $\mathbf{p}_E^B \in \mathbb{R}^3$, which can be obtained through the forward kinematics of the manipulator. The relationship between $\mathbf{p}_E$ and $\mathbf{p}_E^B$ is
\begin{equation} 
\label{eq_pe}
	\mathbf{p}_E = \mathbf{p}_B + \mathbf{R}_B \mathbf{p}_E^B.
\end{equation}

The state space variable is \(\mathbf{\xi} = [\mathbf{p}_B, \mathbf{\Theta}_B, \mathbf{q} ] \in \mathbb{R}^{6+n}\). It can be divided into actuated state variables \(\mathbf{\xi}_c = [\mathbf{p}_B, \psi_B, \mathbf{q} ] \in \mathbb{R}^{4+n}\) and underactuated state variables \(\mathbf{\xi}_u = [\phi_B, \theta_B] \in \mathbb{R}^2\). Differentiating \eqref{eq_pe} yields \cite{chen2023online}:
\begin{align}
\dot{\mathbf{p}}_E = \mathbf{J}\dot{\mathbf{\xi}} =  \mathbf{J}_c\dot{\mathbf{\xi}}_c + \mathbf{J}_u\dot{\mathbf{\xi}}_u,
\end{align}
where $\mathbf{J}_c$ and $\mathbf{J}_u$ are actuated Jacobian matrix and underactuated Jacobian matrix, respectively. We leave the details of the $\mathbf{J}_c$ and $\mathbf{J}_u$ for brevity.  

\subsection{The Modified Kinematic Model}
Ignoring dynamics of the aerial manipulator may result in a reduction in accuracy. To enhance the precision of kinematic coordinate control, a modified kinematic model including the closed-loop dynamics of the aerial manipulator has been proposed. It incorporates two additional sub-models.

The first sub-model is an modified kinematic model for the quadcopter, established using an equivalent model. The equivalent model can effectively replicate the dynamics of complex systems using a low-order model \cite{zou2014survey}. The closed-loop dynamics of the quadcopter are related to its flight controller. According to reference \cite{cao2023eso}, the equivalent model of the quadcopter can be simplified as
\begin{align}
\dot{\p}_B = \mathbf{K}_B(\p_{B,d} - \p_B) + \mathbf{\Delta},
\label{quadcopter_closed}
\end{align}
where \(\p_{B,d} \in \mathbb{R}^3\) represents the desired position of the quadcopter, \(\mathbf{K}_B = {\rm diag}([k_{Bx}, k_{By}, k_{Bz}]) \in \mathbb{R}^{3 \times 3}\) represents the constant parameters of the equivalent model, and \(\mathbf{\Delta} = [\Delta_x, \Delta_y, \Delta_z]^\top \in \mathbb{R}^3\) denotes the linear residual velocity due to modeling errors and uncertainty disturbances. The modified kinematic model of the quadcopter in aerial manipulator is
\begin{align}
\begin{aligned}
\dot{\p}_{B,d} = \mathbf{v}_{B,d}, \dot{\mathbf{v}}_{B,d} = \mathbf{a}_{B,d}, \\
\dot{\p}_B = \mathbf{K}_B(\p_{B,d} - \p_B) + \mathbf{\Delta}.
\end{aligned}
\label{eq_quadcopter_model}
\end{align}

The second sub-model is the kinematic model of the manipulator. The manipulator exhibits high-speed motion and rapid response, its actual values can be considered equivalent to the desired values. The kinematic model of the manipulator is $\dot{\q} = \mathbf{v}_{q,d}, \dot{\mathbf{v}}_{q,d} = \mathbf{a}_{q,d}$ .

Based on the two sub-models, the state space variable of the modified kinematic model can be defined as \(\mathbf{x} = [{p}_{Bx}, \dot{{p}}_{Bx}, {p}_{Bx,d}, \dot{{p}}_{Bx,d}, {p}_{By}\cdots, {q_n}, \dot{{q_n}}]^\top \in \mathbb{R}^{24}\). The residual state variable is defined as \(\mathbf{x}_{res} = [0, \Delta_x, 0_{1 \times 3}, \Delta_y, 0_{1 \times 3}, \Delta_z, 0_{1 \times 14}]^\top \in \mathbb{R}^{24}\). Our control input is \(\mathbf{u} = [\ddot{{p}}_{Bx,d}, \ddot{{p}}_{By,d}, \ddot{{p}}_{Bz,d}, \ddot{{q}}_{1,d}, \cdots, \ddot{{q}}_{n,d}]^\top \in \mathbb{R}^{9}\), which includes the linear accelerations of the quadcopter and the joint accelerations of the manipulator. The control input $\textbf{u}$, after undergoing double integration, serves as the input to the lower level decoupled dynamic controller. The quadcopter utilizes an ESO-based robust controller, while the manipulator employs a PID controller. Our state-space model is
\begin{align}
\mathbf{x}_{k+1} = \mathbf{A} \mathbf{x}_{k} + \mathbf{B} \mathbf{u}_{k} + \mathbf{C} \mathbf{x}_{res,k}, \label{eq_x_trans}
\end{align}
where \(\mathbf{A} = {\rm diag}({\mathbf{A}_{1}}, \mathbf{A}_{1}, \mathbf{A}_{1}, {\mathbf{A}_{2}}, \cdots, \mathbf{A}_{2}) \in \mathbb{R}^{24 \times 24}\), \(\mathbf{B} = {\rm diag}({\mathbf{B}_{1}}, \mathbf{B}_{1}, \mathbf{B}_{1}, {\mathbf{B}_{2}}, \cdots, \mathbf{B}_{2}) \in \mathbb{R}^{24 \times 9}\), and \(\mathbf{C} = {\rm diag}({\mathbf{C}_1}, {\mathbf{C}_1}, {\mathbf{C}_1}, {\mathbf{O}_{12}}) \in \mathbb{R}^{24 \times 24}\). \(\mathbf{O}_{12} \in \mathbb{R}^{12 \times 12}\) are zero matrices. \(\mathbf{B}_{1} = \left[k\delta_t^3/2, k\delta_t^2/2, \delta_t^2/2, \delta_t \right]^\top \in \mathbb{R}^{4}\), where \(\delta_t\) represents the time interval between two consecutive states, \(\delta_t^2\) and \(\delta_t^3\) represent the square and cube of \(\delta_t\),  for the three \(\mathbf{A}_{1}\) and three \(\mathbf{B}_{1}\) matrices, \(k\) takes values \(k_{Bx}\), \(k_{By}\), and \(k_{Bz}\) in sequence. \(\mathbf{B}_{2} = \left[\delta_t^2/2, \delta_t \right]^\top \in \mathbb{R}^{2}\). \(\mathbf{A}_{2} = \left[\begin{array}{cc}1 & \delta_t \\ 0 & 1 \end{array}\right] \in \mathbb{R}^{2 \times 2}\), \(\mathbf{A}_{1} = \begin{bmatrix}1-k\delta_t & 0 & k\delta_t & k\delta^2 \\ -k & 0 & k & k\delta_t \\ 0 & 0 & 1 & \delta_t \\ 0 & 0 & 0 & 1\end{bmatrix} \in \mathbb{R}^{4 \times 4}\) and \(\mathbf{C}_{1} = \begin{bmatrix}0 & \delta_t & 0 & 0 \\ 0 & 1 & 0 & 0 \\ 0 & 0 & 0 & 0 \\ 0 & 0 & 0 & 0 \end{bmatrix} \in \mathbb{R}^{4 \times 4}\).

\subsection{Residual Learning}
In this section, we address the learning of residual \(\mathbf{\Delta} \in \mathbb{R}^{3}\) in the modified kinematic model through neural networks. As shown in  Fig~\ref{frame}, our overall residual learning framework consists of two parts: offline residual learning and online residual learning. 

For offline residual learning, we employ a feedforward neural network with two hidden layers, each with 32 units, and the ELU activation function is used for each hidden layer \cite{saviolo2023active}. The inputs to the neural network include the quadcopter's roll-pitch-yaw Euler angles, linear velocity, angular velocity $\boldsymbol{\omega}_B\in \mathbb{R}^{3}$ and joint angles, joint velocities of the manipulator, and control inputs. The output of the neural network is \(\mathbf{\Delta}\).  We collect data from a 15 cm × 15 cm × 10 cm learning space around the manipulator’s end-effector in its operational configuration (as shown in Fig~\ref{frame}~(b)). During data collection, the end-effector of the manipulator follows a 8-shape trajectory at different speeds in the 3D space. The quadcopter stay its position or traces circular trajectories with a 10 cm radius at different speeds in the 3D space. Each stage lasts 150 seconds to collect suffcient and informative data.

Online residual learning adjusts the parameters of the final layer of the network in real-time. We define the model prediction error based on the deviation between the predicted state and the actual measured state of the modified kinematic model as \(\mathcal{L}_k = \| \hat{\dot{\mathbf{p}}}_{B,k} - \dot{\mathbf{p}}_{B,k} \|^2\), where \(\mathcal{L}_k\) is the model prediction error at time step \(k\),  $\hat{\dot{\mathbf{p}}}_{B,k}$ is the predicted velocity of the learning-based modified kinematic mode and \(\dot{\mathbf{p}}_{B,k}\) represent the measured linear velocity of the quadcopter at time step \(k\), respectively. We use stochastic gradient descent (SGD) to adjust the weights of the last layer of the neural network, specifically \cite{saviolo2023active}:
\begin{align}
    \theta_k^\ell = \theta_{k-1}^\ell - \eta \frac{1}{B_l} \sum_{i=k-B_l}^k \nabla \mathcal{L}_i,
    \label{eq_online}
\end{align}
where \(\theta^\ell_k\) and \(\theta^\ell_{k-1}\) represent the parameters of the last layer of the neural network at time steps \(k\) and \(k-1\), respectively, \(\eta\) is the learning rate, and \(B_l\) is the batch size. By updating the network parameters using multiple sets of data, online residual learning can be made more stable.
\section{Model Predictive Control Based on Weight Allocation}

This section introduces the proposed MPC framework based on weight allocation. By coordinating the distribution of the weight matrices in the MPC optimization, the framework achieves kinematic coordinate control of the aerial manipulator and facilitates the switching between different motion modes.
\subsection{MPC formulation}
The MPC framework formulates the kinematic coordinate control problem as a constrained optimization problem, ensuring that the aerial manipulator strictly adheres to physical constraints and maintains safety during operation. Furthermore, by predicting the position of the quadcopter, the manipulator can be pre-planned to compensate for end-effector deviations caused by discrepancies between the actual and desired states of the quadcopter, thereby enhancing the precision of end-effector kinematic control. Our specific MPC formulation is as follows:
\begin{align}
\begin{aligned}
    &\hspace{32pt} \min_u \sum_{k=0}^N J(\mathbf{x}_k, \mathbf{u}_k, \mathbf{x}_{res,k}), \\
    &s.t. \hspace{8pt} \mathbf{x}_{k+1} = \mathbf{A} \mathbf{x}_k + \mathbf{B} \mathbf{u}_k + \mathbf{C} \mathbf{x}_{res,k}, \\
    &\hspace{24pt} \mathbf{x}_k \in \mathcal{X}, \mathbf{u}_k \in \mathcal{U}, \\
    &\hspace{24pt} \mathbf{x}_0 = \mathbf{x}_{init},
 \label{eq_mpc}
\end{aligned}
\end{align}
where \eqref{eq_x_trans} defines our state-space kinematic model, and \(\mathbf{x}_{init}\) represents the initial state. \(\mathcal{X}\) and \(\mathcal{U}\) are the boundaries of the state and control variables, respectively. \(N\) is the prediction horizon of the MPC, and \(J(\mathbf{x}_k, \mathbf{u}_k, \mathbf{x}_{res,k})\) is our objective function.

\subsection{Objective Function}

To achieve high-precision manipulation of the aerial manipulator’s end-effector and kinematic coordinate control, the objective function consists of four parts, expressed as
\begin{align}
J(\mathbf{x}_k, \mathbf{u}_k, \mathbf{x}_{res,k}) = J_1 + J_2 + J_3 + J_4.
\label{eq_costf}
\end{align}

The first part aims to minimize the end-effector tracking error. Specifically, \({J}_1 = \|\mathbf{J}_c\dot{\mathbf{\xi}}_c + \mathbf{J}_{u}\dot{\mathbf{\xi}}_{u} - \dot{\mathbf{p}}_{E,d}\|_{\mathbf{W}_1}^2\), where \(\mathbf{W}_1\in \mathbb{R}^{3}\) is a diagonal weight matrix. The desired end-effector velocity can be defined as \(\dot{\mathbf{p}}_{E,d} = \dot{\mathbf{p}}_{E,c} - \mathbf{K}_e \mathbf{e}_E\) \cite{cao2020predictive,faroni2018predictive}, \(\mathbf{K}_e \in \mathbb{R}^{3 \times 3}\) is a diagonal constant matrix, \(\dot{\mathbf{p}}_{E,c} \in \mathbb{R}^{3}\) is the velocity of the target trajectory, and \(\mathbf{e}_E = \mathbf{p}_E - \mathbf{p}_{E,d}\) is the end-effector postion error, with \(\mathbf{p}_E \in \mathbb{R}^{3}\) and \(\mathbf{p}_{E,d} \in \mathbb{R}^{3}\) representing the current position and the desired position from the target trajectory.

The second part aims to minimize the linear velocity of the quadcopter and joint velocity of the manipulator. Specifically, \({J}_2 = \|\mathbf{v}_s\|_{\mathbf{W}_2}^2\), where \(\mathbf{v}_s = [\dot{p}_{Bx}, \dot{p}_{By}, \dot{p}_{Bz}, \dot{q}_1, \cdots, \dot{q}_n]^\top \in \mathbb{R}^9\) and \(\mathbf{W}_2\in \mathbb{R}^{9}\) is a diagonal weight matrix. Optimizing \(\mathbf{v}_s\) can achieve coordinated motion allocation between the quadcopter and the manipulator.

The third part aims to minimize the control inputs, defined as \({J}_3 = \|\mathbf{u}\|_{\mathbf{W}_3}^2\), where \(\mathbf{W}_3\in \mathbb{R}^{9}\) is a diagonal weight matrix. Minimizing the control inputs helps to avoid drastic changes in the control signals, conserving energy and extending the manipulation time of aerial manipulators.

The fourth part ensures that during manipulation, the end-effector remains within the model's learning space. Specifically, \({J}_4 = \|\p_{E}^B - \p_{O}^B\|_{\mathbf{W}_4}^2\), \(\p_{O}^B\) is the position of the center of the learning space in the quadcopter’s body frame, and \(\mathbf{W}_4\in \mathbb{R}^{3}\) is a diagonal weight matrix. It can ensuring high manipulability and precision while preventing excessive manipulator movements that could cause significant quadcopter instability. 

\subsection{Weight Allocation}

\begin{figure}
\centering
\includegraphics[width=0.9\linewidth]{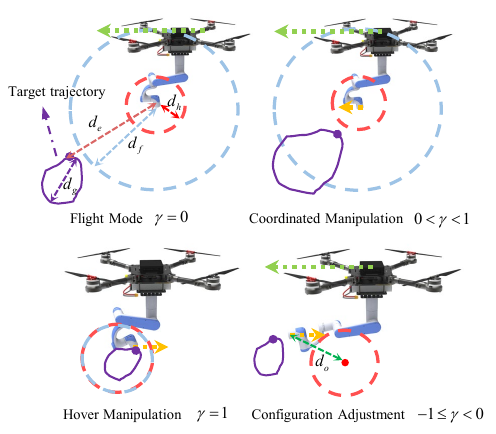}
\caption{\label{fig_sys}
 Different mode of the Aerial Manipulator. The red and blue circles represent State Transition Boundaries.}
\end{figure}

Our allocation strategy consists of three steps: first, the desired motion state of the manipulator is evaluated; second, the weights are adjusted based on the evaluation; finally, the MPC controls the aerial manipulator according to the adjusted weights. To better evaluate the desired motion state of the manipulator, we define \(\gamma \in [-1,1]\) as a metric for the expected motion, Specifically:
\begin{align}
\gamma = \begin{cases} 
0 & \text{if } d_e > d_f, \\
\frac{k_{mp} + d_h}{k_{mp} + d_e} & \text{if } d_h < d_e \leq d_f \ and ~d_o < d_h, \\
1 & \text{if } d_e \leq d_h,   d_o<d_h \ and ~ d_g < d_h, \\
-k_{mn}\frac{d_o - d_h} {d_{edge} - d_h} & \text{if } d_o \geq d_h,
\end{cases}
\label{eq_gama}
\end{align}

where \(d_e\) is the distance between the end-effector and the target trajectory (as shown in Fig~\ref{fig_sys}.); \(d_f\) is the boundary for switching between flight mode and manipulation mode; \(d_h\) is the boundary for switching between the hover manipulation mode and other modes and it is related to the boundary of the learning space; \(d_o\) represents the distance from the end-effector to the center of the learning space; \(d_g\) represents the maximum diameter of the target trajectory; \(d_{edge}\) represents the maximum distance from the center of the manipulation space to the edge of the manipulator’s workspace; \(k_{mp}\) and \(k_{mn}\) are constants greater than 0.
The value of \(d_f\) depends on two cases:  when the target trajectory cannot be fully covered by the hover manipulation space, \(d_f\) is a large fixed boundary, and the system enters coordinated manipulation mode ( \(0 < \gamma < 1\)) after the flight approaches the target trajectory; when the size of the target trajectory \(d_g\) is smaller than the hover manipulation boundary, \(d_f = d_h\), it enters the hover manipulation mode (\(\gamma = 1\)) after the flight mode approaches the target. When \(d_e\) is large, the system remains in flight mode, where \(\gamma = 0\). \(d_o\) is used to assess whether the manipulator end-effector is within the learning space. If it exceeds the workspace, \(-1 \leq \gamma < 0\). In configuration adjustment mode, the end-effector of the manipulator moves towards the learning space, while the quadcopter moves towards the target trajectory to ensure the trajectory tracking performance of the aerial manipulator end-effector.

\begin{figure*}[t]
\centering
\includegraphics[width=0.95\linewidth]{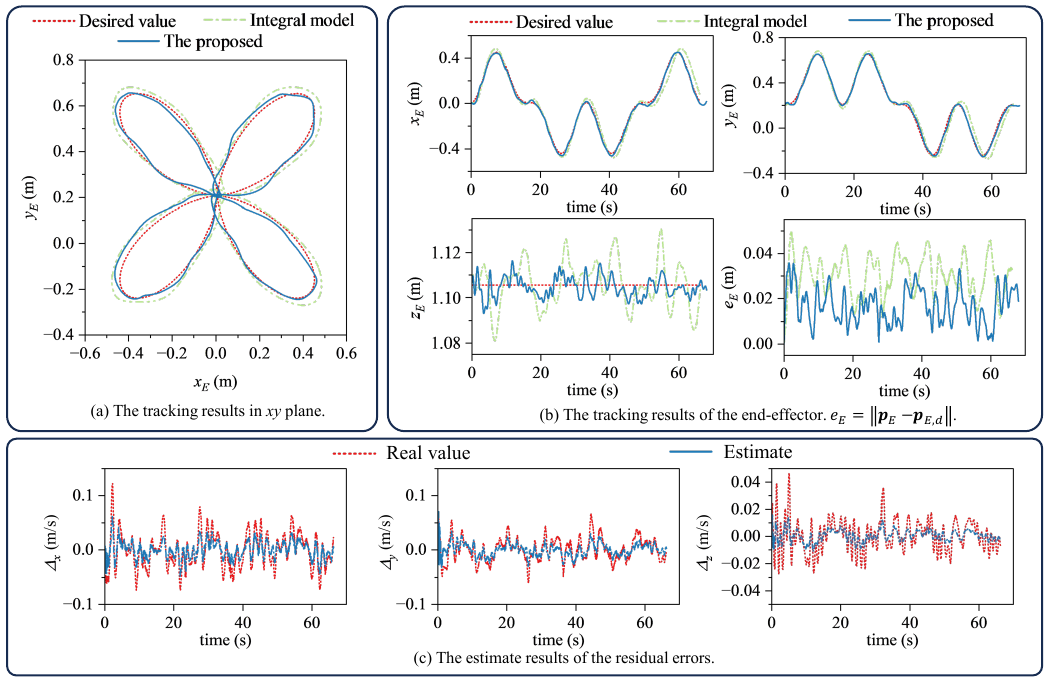}
\caption{\label{result_1}The results of the complex trajectory tracking experiment.}
\end{figure*}

The second part involves adjusting the weight matrices in the MPC objective function. Adjusting \(\mathbf{W}_2\) changes the velocity of the quadcopter and manipulator, while adjusting \(\mathbf{W}_4\) ensures that the manipulator stays close to the learning space. Specifically, \(\mathbf{W}_2 = \mathbf{A}_2 \mathbf{W}_{2,0}\) and \(\mathbf{W}_4 = w_d \mathbf{W}_{4,0}\), where \(\mathbf{W}_{2,0}\) and \(\mathbf{W}_{4,0}\) are the initial weight matrices. \(\mathbf{A}_2 = {\rm diag}([w_q, w_q, w_q, w_m, \cdots, w_m]) \in \mathbb{R}^{9 \times 9}\), where \(w_q\) and \(w_m\) are parameters adjust the velocity of the quadcopter and the manipulator, respectively, and the parameter \(w_d\) adjusting the weight in the objective function to bring the manipulator’s end-effector closer to the center of the learning space. The specific definitions are
\begin{align}
w_q = k_q(\gamma^2 + 0.1), w_m = \frac{k_m}{\gamma + 0.01}, w_d = \frac{k_d d_o}{1.1 + \gamma},
\label{eq_w}
\end{align}
where \(k_q\), \(k_m\), and \(k_d\) are constants greater than 0. As \(|\gamma|\) increases, we decrease \(w_m\) and increase \(w_q\), thereby increasing the movement weight of the manipulator in the MPC while reducing the movement weight of the quadcopter.
\def\SetClass{article}
\begin{algorithm}
\caption{MPC based on weight allocation}\label{algorithm}
$\mathbf{Input}$: $d_{g}$, $\mathbf{p}_{E}$,  $\mathbf{p}_{E,d}$, $\mathbf{x}_{k}$, $d_e$,$d_o$, \(\theta^\ell_{k-1}\)\;
$\mathbf{Output}$: $\mathbf{a}_{B,d}$, $\ddot{\mathbf{q}}_d$\;
$ \hspace {16pt}\mathbf{function}$  MPC($d_{g}$, $\mathbf{p}_{E}$,  $\mathbf{p}_{E,d}$, $\mathbf{x}_{k}$)\;
$ \hspace {32pt}\gamma  \leftarrow   $  calculating  \eqref{eq_gama}\;
$\hspace {32pt}w_q$, $w_m$, $w_d  \leftarrow$ calculating  \eqref{eq_w}\;
$\hspace {32pt}$$\mathbf{J}_{}\leftarrow$ calculating   \eqref{eq_costf}\;
$\hspace {32pt}$NN residual model update  $\leftarrow$ calculating  \eqref{eq_online}\;
$\hspace {32pt}$update modified kinematic model using  \eqref{eq_quadcopter_model}\;
 $\hspace {32pt}\mathbf{a}_{B,d}$, $\ddot{\mathbf{q}}_d  \leftarrow   $  calculating   \eqref{eq_mpc}\;
$\hspace {32pt}\mathbf{return}$  $\mathbf{a}_{B,d}$,$\ddot{\mathbf{q}}_d$\;
$ \hspace {16pt}\mathbf{end}~\mathbf{function}$
\end{algorithm}

\section{Experimental Validation}
In this section, we validate the proposed method by two numerical experiments.
\begin{figure*}[t]
\centering
\includegraphics[width=0.95\linewidth]{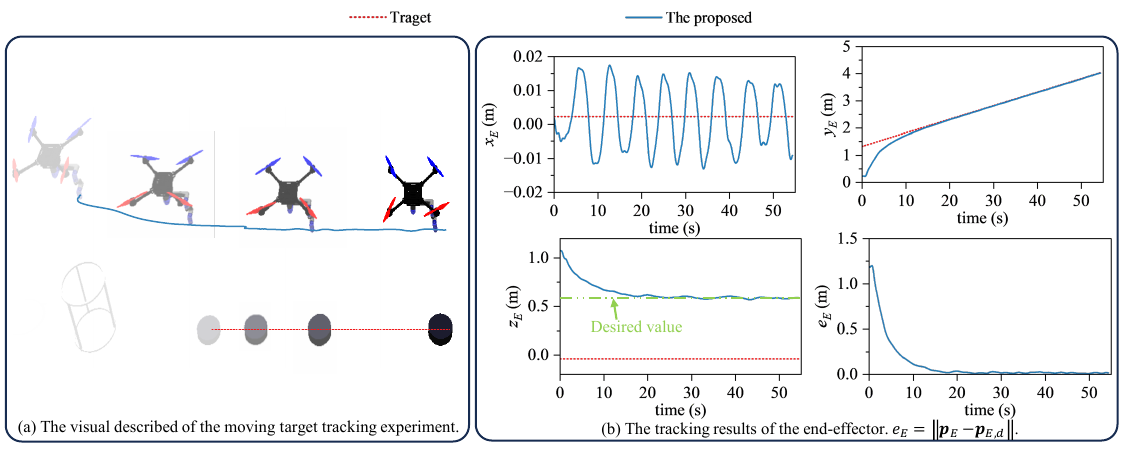}
\caption{\label{result_2}The results of the moving target tracking experiment.}
\end{figure*}

\subsection{Experimental Setup}
The quadcopter used in this study has a diagonal wheelbase of 0.93 m, with a length and width of 1.37 m and a height of 0.12 m. Its mass is 2.6 kg. The manipulator is mounted beneath the quadcopter, with a total mass of 1.2 kg. We validated our algorithm in Gazebo. The quadcopter uses a ESO-based robust controller, running in a PX4 environment at a frequency of 100 Hz. The detials can be found in our previous work \cite{cao2023eso}. The neural network of offline residual learning is implemented in Pytorch, while the online residual learning updates the parameters in C++. The MPC is implemented through the acados \cite{verschueren2022acados} framework with a horizon $N$ = 15 and planning frequency of 50 Hz. In all simulations, we used the same control gains, with the modified kinematic parameters set to \(\mathbf{K}_{B} = {\rm diag}([6.67, 6.67, 2.38])\). The learning rates \(\eta\) were set to 0.01 for offline residual learning and 0.0015 for online residual learning, with a batch size \(B_l = 20\). The matrix \(\mathbf{K}_{e} = {\rm diag}([0.8, 1.2, 1.2])\), the values of \(k_q\) was set to 1000, \(k_m\) was set to 10 and \( k_d\)  was set to 100, the values of \(k_{mn}, k_{mp}\) were set to 1. The parameters \(d_{edge}, d_f, d_h\) were set to 0.38 m, 1.0 m, and 0.075 m, respectively.

\subsection{Complex Trajectory Tracking Experiment}
The purpose of this experiment is to verify that the end-effector can track complex trajectories under the proposed algorithm framework. The desired trajectory is a four-leaf clover pattern. The desired trajectory of the end-effector is given in Fig~\ref{result_1}(a).

The tracking performance of the end-effector is shown in Fig~\ref{result_1}, and the integral model is a model that does not consider dynamic characteristics. From Fig~\ref{result_1}~(b), the average position error of the proposed method is 0.0122 m, compared to 0.0302 m for MPC using the integral model, the proposed method improved accuracy by 59.6\%, which demonstrates the effectiveness of our learning-based modified kinematic model. Fig~\ref{result_1}~(c) is the estimate results of residual errors, it indicates that our neural network can accurately estimate the trend of changes in uncertainty disturbances.

\subsection{Moving Target Tracking Experiment}
The purpose of this experiment is to verify that the aerial manipulator can achieve coordinated motion across different modes under the proposed algorithm framework. The initial coordinates of the aerial manipulator end-effector and the moving target are \([0, 0.2, 1.2]^T\) m and \([0, 1.4, 0]^T\) m, respectively. The moving target moves along the positive y-axis at a speed of 0.05 m/s. The aerial manipulator is required to start from the hover position, catch up with the moving target, and follow it at a height of 0.6 m above the target.

As shown in Fig~\ref{result_2}, the aerial manipulator catches up with the target at around 15 seconds and follows it smoothly. The trajectory throughout the process is relatively smooth, indicating that the aerial manipulator can stably switch from hover mode to flight mode and then to coordinated mode. The manipulator approaches the object within 5 seconds and enters a stable tracking state after 15 seconds. The average error during stable end-effector tracking is 0.0197 m. This demonstrates the high-precision kinematic coordinate control across different modes, verifying the effectiveness of our  kinematic coordinate control method.

\section{CONCLUSIONS}
This paper proposes a predictive kinematic coordinate control method, which includes a learning based modified kinematic model and a MPC scheme based on weight allocation. The method is validated through two experimental results. The first experiment shows that, compared to MPC using the integral model, the proposed method reduces the average error by 59.6$\%$, verifying the effectiveness of our learning-based modified kinematic model for high-precision control. The second experiment demonstrates that our aerial manipulator can quickly approach the target and stably follow its motion, proving the effectiveness of our motion allocation strategy. Although we have achieved improvements in trajectory accuracy and coordinated tracking of moving targets, validating the algorithm on a physical platform rather than in simulation would provide stronger conclusions. Our residual learning model can effectively predict the residuals and ensure that MPC achieves a control frequency of 50 Hz, there is still room for improvement in residual learning. In the future, we aim to enhance the residual learning capability while maintaining the real-time performance of MPC.


\end{document}